\documentclass{article}
\usepackage{iclr2027_conference,times}
\usepackage{hyperref}
\usepackage{url}
\usepackage{booktabs}
\usepackage{graphicx}
\usepackage{amsmath}
\usepackage{amssymb}
\usepackage[table]{xcolor}
\usepackage{tcolorbox}
\usepackage{float}
\usepackage{dblfloatfix}
\usepackage{bm}
\usepackage{alltt}
\title{TACO-Graph: Automated Construction \\ of Text-Attributed Graphs}

\author{Amiri Hayes \\ MIT CSAIL \\ \texttt{amiri@mit.edu} \And Kristina Wicke \\ New Jersey Institute of Technology \\ \texttt{kristina.wicke@njit.edu}}

\iclrfinalcopy
\begin{document}

\maketitle
\lhead{Preprint}
\begin{abstract}
Graph-structured representations of text data have become 
foundational across machine learning, yet it remains unclear how to systematically construct a graph from a given dataset for a target task. Choices about which entities become nodes, which relationships become edges, and what text each node contains can substantially affect downstream performance. In practice, finding optimal constructions often requires expert knowledge or manual trial and error. We introduce TACO-Graph, a decision tree framework for learning which graph construction choices are most effective for a given dataset and task. Specifically, we focus on text-attributed graphs (TAGs) constructed from multi-attribute tabular data. TACO-Graph enumerates valid TAG representations, scores each using fast proxy evaluation on small representative subsets, and trains an interpretable decision tree on those scores. Across five datasets spanning citation networks, product reviews, and bibliographic relationships, our decision trees predict construction quality on separately drawn evaluation samples with mean Spearman $\rho$ above $0.9$ for node classification and above $0.8$ for GraphRAG retrieval. Our trees systematically identify which constructions perform better across both GNN and GraphRAG tasks, and the full pipeline is released as an open-source library to support evaluation on new tasks and datasets at \url{https://github.com/AmiriHayes/text_attributed_graphs}. TACO-Graph provides a systematic and interpretable guide for determining how to convert tabular data to text-attributed graphs for downstream tasks.
\end{abstract}

\section{Introduction}
\label{sec:intro}
Text-Attributed Graphs (TAGs) increasingly serve as the basis for a range of applications, including graph neural network (GNN) inference~\citep{hamilton2017, velickovic2018}, Graph Retrieval-Augmented Generation (GraphRAG) retrieval systems~\citep{edge2024, guo2024lightrag}, and knowledge graph question answering~\citep{wang2024kgprompt, peng2024graphrag}. However, existing TAG-based methods typically assume that the graph construction is given or fixed, without systematically considering how choices of nodes, edges, and textual attributes affect downstream performance~\citep{tape2023, zhao2023glem, liu2024ofa, cstag2023}. For example, the same product review data can be structured as a graph of products linked by co-purchased items or a graph of users linked by shared purchase histories. These choices yield substantially different graph topologies and text distributions, directly altering downstream model performance, yet we lack a systematic understanding of which construction will succeed on a given dataset and task.
A similar pattern holds in relational databases, where construction relies on practitioner intuition \citep{gan2024multitable}, even though optimal schema-level choices are shown to differ in performance by up to 10\% in \citet{rdb2gbench2025}. AutoG in \citet{autog2025} introduce an LLM agent that proposes and evaluates schema
transformations for relational databases. However, AutoG neither formalizes a general construction space nor generates interpretable, task-conditioned construction rules.

\begin{figure}[htbp]
    \centering
    \includegraphics[width=\textwidth]{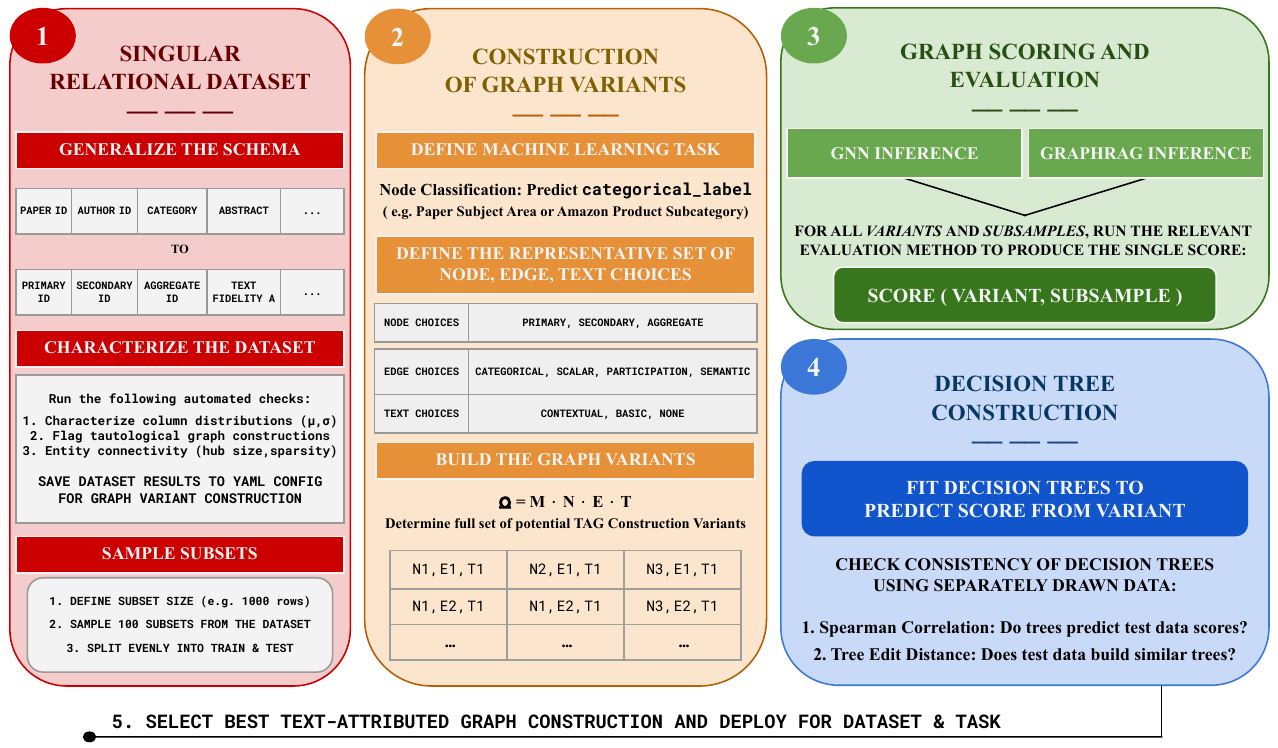}
    \caption{\textbf{Overview of TACO-Graph.}
    \textbf{(1)} A relational dataset's columns are mapped to a generalized schema and characterized using automated statistical and validity checks. These dataset level properties are stored for each dataset, and train/test subsets are sampled. \textbf{(2)} For a specified downstream task, TACO-Graph enumerates the space of potential TAG constructions (variants) by defining node, edge, and text-fidelity choices. \textbf{(3)} Each construction is evaluated on representative subsamples by scoring (via lightweight GNN or small-sample GraphRAG), yielding $\mathrm{Score}(\mathrm{variant}, \mathrm{subsample})$. \textbf{(4)} A decision tree is fit to variant construction scores and validated on separately drawn data using Spearman correlation and tree edit distance. \textbf{(5)} The interpretable ranking from the tree reveals optimal and suboptimal graph construction choices, informing the practitioner.}
    \label{fig:centerpiece}
\end{figure}

We propose TACO-Graph (Text-Attributed Construction Optimizer for Graphs), a predictive framework that learns effective TAG constructions for a given dataset and task (Figure~\ref{fig:centerpiece}). Rather than selecting constructions ad hoc, TACO-Graph systematically enumerates valid choices and evaluates these choices on small, representative data subsets. These evaluations train interpretable decision trees to rank constructions without full-dataset evaluation. We evaluate TACO-Graph across five datasets spanning three domains, focusing on building graphs for node classification and GraphRAG retrieval. Our results show that trees trained on small subsets accurately predict performance on separately drawn data, with reliable rankings emerging from small sample budgets across both GNN classification and GraphRAG retrieval, though optimal constructions differ by task.

\section{Related Work}
\label{sec:related}
\subsection{Learning from Text-Attributed Graphs}
Inference on text-attributed graphs spans multiple axes: the textual content of nodes encodes semantic meaning, node types determine what entities are represented, and edge structure defines the relationships. TAGs have been used across a range of applications, including academic research~\citep{clement2019arxiv}, e-commerce~\citep{mcauley2023}, and biomedical discovery~\citep{hu2020ogb}. Prior work examines both TAG construction and learning: TAPE~\citep{tape2023}, for example, demonstrates that different LLM-generated node features can lead to substantially different downstream GNN performance on citation graphs, while CS-TAG~\citep{cstag2023} provides standardized benchmarks for text-attributed graph learning. GNN architectures including GraphSAGE~\citep{hamilton2017} and GAT~\citep{velickovic2018} provide common approaches for learning over TAGs, while more recent methods such as GLEM~\citep{zhao2023glem} and OFA~\citep{liu2024ofa} explore tighter integration between language model pre-training and GNN fine-tuning. Across these approaches, the graph is given: node entities, edge relationships and textual attributes are fixed before learning begins rather than treated as variables to optimize. A recent line of work has begun to question this assumption for relational data~\citep{autog2025, autograble2026, relatron2026, schemagraphs2026}, which we discuss next.

Recent work takes first steps toward automating graph construction in relational settings. AutoG~\citep{autog2025} uses LLM agents to propose schema transformations for relational databases, and AutoGrable~\citep{autograble2026} scores constructions via label alignment without training GNNs, though it fixes rows as nodes, does not consider text, and functions as a screen rather than a ranking for downstream performance. Adjacent structural work learns schema graphs~\citep{schemagraphs2026}, decomposes edge types~\citep{rose2024}, and optimizes architectures over fixed graphs~\citep{relatron2026}. TACO-Graph instead treats node entities, edge relationships, and textual attributes as explicit optimization variables, learning effective combinations for a dataset and task.

\subsection{Construction of Text-Attributed Graphs}
The idea of searching over structured design spaces using surrogate models is well-established. Neural Architecture Search~\citep{elsken2019nas} automates architectural component selection by learning from observed performance over a structured search space, while AutoML pipelines extend this approach to hyperparameter and feature selection~\citep{hutter2019automl}. Graph structure learning methods, including Learning Discrete Structures (LDS) ~\citet{franceschi2020} and ProGNN~\citep{jin2020prognn}, instead learn graph adjacency jointly with GNN parameters. These approaches share with TACO-Graph the idea of using inexpensive evaluations to guide search, but are not specifically designed to determine how tabular data should be represented as a text-attributed graph. We similarly treat construction choices as categorical variables, evaluate sampled configurations using fast metrics, and fit an interpretable predictive model to identify high-performing constructions.

Graph-based retrieval-augmented generation (GraphRAG) is a new inference paradigm that organizes a retrieval corpus as a graph, enabling structured traversal alongside vector retrieval. \citet{edge2024} introduced community-based GraphRAG, which applies Leiden community detection and retrieves at the community level for global summarization queries. LightRAG~\citep{guo2024lightrag} combines local entity-level and global graph-level retrieval, while HippoRAG~\citep{gutierrez2024hipporag} and G-Retriever~\citep{he2024gretriever} explore graph-structured retrieval for multi-hop question answering. Despite this diversity, GraphRAG systems typically rely on fixed or manually designed graph-construction procedures, leaving the effect of alternative TAG constructions largely unexplored. To address this gap, we investigate whether small-subset evaluation can identify effective constructions prior to full-scale GraphRAG construction. 

\section{Methodology}
\label{sec:method}

\begin{table}[h]
\centering
\caption{TAG construction search space \textbf{$\Omega_{\mathcal{M}}$} with \textit{citation network examples}.} \vspace{5pt}
\label{tab:ontology}
\small
\setlength{\tabcolsep}{6pt}
\linespread{1.2}\selectfont
\renewcommand{\arraystretch}{1.2}
\begin{tabular}{|p{0.45\columnwidth}|p{0.49\columnwidth}|}
\hline
\textit{Tasks (M)} & \textit{Nodes (N)} \\
\hline
\cellcolor{gray!15}
1. \textbf{Node Classification}: predict the  category labels of nodes. | \textbf{\textit{e.g. paper subject classification}} \phantom{text}\newline 
2. \textbf{GraphRAG Retrieval}: answer questions over graph-structured context. | \textbf{\textit{e.g. author queries}} \phantom{text}
&
\cellcolor{gray!15}
\textbf{N1} Primary Entity: the fundamental unit of 
information with core textual content. | \textbf{\textit{e.g. Paper}}\newline
\textbf{N2} Secondary Entity: hub node connecting 
primary entities through shared participation. | \textbf{\textit{e.g. Author}} \newline
\textbf{N3} Aggregate Entity: a higher-order grouping 
of primary entities by category or family. | \textbf{\textit{e.g. Journal}}
\\
\hline
\textit{Edges (E)} & \textit{Text (T)} \\
\hline
\cellcolor{gray!15}%
\textit{Ground Truth Metrics:} \newline
\textbf{E4a} Categorical edges | \textbf{\textit{e.g. same category}} \newline
\textbf{E4b} Scalar edges | \textbf{\textit{e.g. similar citation count}} \newline
\textbf{E4c} Participation edges | \textbf{\textit{e.g. coauthorship}} \newline
\textit{Calculated Similarity Metrics:} \newline
\textbf{E5a} Semantic edges | \textbf{\textit{e.g. similar abstract}} \newline
\textbf{E5b} Structural edges | \textbf{\textit{e.g. similar degree}}
&
\cellcolor{gray!15}
\textbf{T6a} Contextual: node text including both primary 
content and its surrounding semantic context \newline| \textbf{\textit{e.g. title + abstract}} \newline
\textbf{T6b} Standard: primary content only with no surrounding context | \textbf{\textit{e.g. title}} \newline
\textbf{T6e} Baseline: no text features, serving 
as a structure-only control. | \textbf{\textit{e.g. no text}}
\\
\hline
\end{tabular}
\end{table}

Our approach begins by explicitly characterizing the space of possible text-attributed graph constructions. A practitioner constructing a TAG from a tabular dataset must make three types of choices: which entities to represent as nodes, which relationships to represent as edges, and which textual attributes to associate with each node. We organize these choices into finite sets of principled alternatives, denoted by $\mathcal{N}$, $\mathcal{E}$, and $\mathcal{T}$, respectively, as summarized in Table~\ref{tab:ontology}. We additionally consider $\mathcal{M}$, the set of downstream machine learning tasks for which we optimize our TAG construction. The resulting space of construction configurations is the Cartesian product $\Omega_{\mathcal{M}} = \mathcal{N} \times \mathcal{E} \times \mathcal{T}$ and encompasses every valid combination of node entity, edge connection, and text fidelity types.

\subsection{Defining TAG Construction Space}
\label{sec:ontology}
The node axis $\mathcal{N}$ specifies which entities are represented as nodes. Primary entities correspond to fine-grained dataset units (e.g., papers in citation networks, reviews in product-review networks), secondary entities serve as hubs connecting primary entities (e.g., paper authors, product reviewers), and aggregate entities represent higher-order groupings (e.g., journals, product families). The edge axis $\mathcal{E}$ specifies how nodes are connected, either through observed relationships in the underlying data (`ground truth metrics') or through computed similarity measures. Relationship-based edges are derived directly from observed relationships, such as co-authorship or co-purchase links, while similarity-based edges are constructed using similarity between node representations. Finally, the text-fidelity axis $\mathcal{T}$ specifies the textual information associated with nodes. Motivated by the observation that different content units carry different informational value~\citep{nenkova2004pyramid}, we define three fidelity levels: full contextual text, primary content only, and no text.

Our ontology defines a representative construction space rather than exhaustively enumerating every possible graph construction. In particular, some construction categories admit many possible parameterizations; for example, infinitely many similarity functions could be used to instantiate E5a. We therefore select representative parameterizations for each construction choice, as detailed in Appendix~\ref{app:datasets}. Conversely, not every combination of construction choices in $\Omega_{\mathcal{M}}$ is valid or informative. We apply three logical pruning constraints before training and evaluation: (1) the selected node entity must possess ground-truth task labels (valid supervision signal), (2) the edge construction must not yield degenerate topologies (e.g., empty or near-complete graphs), and (3) edge rules must not cause label leakage by directly encoding target labels (e.g., pairing aggregate nodes N3 with label edges E4a). These constraints are saved and enforced programmatically before scoring. We further exclude empirically degenerate variants for which more than 95\% of train-pool GNN scores are zero. Together, these constraints remove constructions that are invalid, structurally degenerate, or uninformative while preserving the broader diversity of the construction space.

\subsection{Graph Scoring and Evaluation}
\label{sec:scoring}
We study TAG construction in two downstream settings (GNN-based node classification and GraphRAG question answering) as well as across five initially tabular datasets (ArXiv~\citep{clement2019arxiv}, Sporting Goods, Electronics, and Toys from Amazon Reviews ~\citep{mcauley2023}, and Books-History~\citep{cstag2023}). After preprocessing the datasets into the common schema defined in Table \ref{tab:ontology}, we draw samples for the tasks with the goal of building graph variants on samples and scoring them (Table \ref{tab:scoring_configs}). The end product of scoring is a construction-performance table: for each dataset, subset, and variant in $\Omega_\mathcal{M}$ their recorded scores are used as training data for a decision tree. 

\begin{table}[H]
\centering
\small
\caption{Fixed Scoring configurations.}
\vspace{5pt}
\label{tab:scoring_configs}
\begin{minipage}[t]{0.45\textwidth}
\centering
\textbf{(a) GNN Scoring} \\[3pt]
\begin{tabular}{@{}p{0.38\linewidth} p{0.58\linewidth}@{}}
\toprule
Architecture         & 2-layer GraphSAGE \\
Aggregation          & Mean \\
Node features        & SBERT all-MiniLM-L6-v2 (384-d) \\
Hidden dim           & 256 \\
Output dim           & Dataset- and task-specific \\
Epochs               & 100 \\
Early stopping       & Patience 30 (on validation) \\
Optimizer            & Adam, lr $= 0.001$ \\
Train/val/test split & 70\% / 15\% / 15\% of subset nodes \\
\bottomrule
\end{tabular}
\end{minipage}%
\hfill
\begin{minipage}[t]{0.50\textwidth}
\centering
\textbf{(b) RAGAS Scoring} \\[3pt]
\begin{tabular}{@{}p{0.32\linewidth} p{0.64\linewidth}@{}}
\toprule
Generator / Judge    & GPT-4o-mini, temperature $0.0$ \\
Index                & LlamaIndex VectorStoreIndex \\
Embeddings           & OpenAI text-embedding-3-small \\
Retrieval            & Top-$k$ dense, $k = 3$ \\
Expansion            & 1-hop graph neighbors, capped at 15 passages \\
Metrics              & RAGAS: faithfulness, answer relevance, context relevance; composite $=$ mean \\
Questions            & 300 per dataset \\
Train/test split     & 150 / 150 \\
\bottomrule
\end{tabular}
\end{minipage}
\end{table}

\subsubsection{GNN Scoring for Node Classification}

For each dataset, we draw 150 random 1{,}000-row subsets (split evenly into train and test) from pools of 20k--50k rows, so each run sees $2.0\%$--$4.8\%$ of its pool. Subsets must only preserve relative construction rankings rather than full data distributions. On each subset, we instantiate all valid TAG constructions in $\Omega_{\mathcal{M}}$ and train a GNN on each construction variant up to 100 epochs with early stopping. Each construction is scored on its separately drawn test nodes by a floored McFadden
pseudo-$R^2$ \citep{mcfadden1974},
$S = \max\!\left(0,\, 1 - \mathrm{LL}_{\text{model}}/\mathrm{LL}_{\text{null}}\right)$,
which is a calibration-sensitive score rather than accuracy, and the floor means variants performing below the class prior
are not ranked against one another. The GNN architecture and hyperparameters (Table~\ref{tab:scoring_configs}a) are held fixed across all runs so performance differences only reflect construction choices. 

\subsubsection{RAGAS Scoring for Retrieval Augmented Generation}
For each dataset, we generate 300 evaluation questions using GPT-4o-mini, stratified evenly across three question types: single-document details (local precision), cross-document aggregation (global details), and multi-hop entity traversal (structural reasoning). Retrieval Augmented Generation Assessment (RAGAS)~\citep{ragas2023} provides automated evaluation of retrieval-augmented generation by measuring whether an LLM response is supported by retrieved context (faithfulness), whether the retrieved context is relevant to the question (context relevance), and whether the answer addresses the question (answer relevance). Questions must pass a RAGAS faithfulness threshold ($> 0.5$) before inclusion, yielding a 150/150 train-test split per dataset, except History, where only 233 questions passed and the split is 116/117. We evaluate all valid variants in $\Omega_{\mathcal{M}}$. For each variant, we construct a text-based GraphRAG system over the pooled corpus via LlamaIndex, where retrieved nodes are expanded along the variant's edges to form the answering context. We describe this in Appendix~\ref{app:datasets}. Retrieval quality is scored using the RAGAS composite score (mean of Faithfulness, Answer Relevance, and Context Relevance across questions). As in the GNN setting, scores serve as targets for the decision-tree model, measuring construction quality via retrieval performance.

\subsection{Decision Tree Training and Validation}
\label{sec:dttree}

From construction performance tables, we train a decision tree regressor using per-variant train-pool mean scores as targets. We encode each construction tuple $(N, E, T)$ as a categorical feature vector, training the decision tree to map design choices directly to predicted downstream performance. To test whether these rules generalize, we evaluate two complementary metrics: (1) Spearman's rank correlation $\rho$ between tree predictions and test-pool scores, measuring whether the tree preserves the relative ordering of construction quality, and (2) tree edit distance (TED) between train-pool and test-pool trees, measuring structural rather than predictive agreement. While \citet{relatron2026} warn that validation performance can be an unreliable architecture selector due to temporal splits in relational tasks, our pools are randomly drawn from the same dataset without temporal drift, though we note they are drawn independently rather than as a strict partition. We separately validate that this does not affect our results by repeating the analysis on strictly disjoint pools (Section~\ref{sec:results_graphrag}). Together, high $\rho$ and low TED indicate that the tree learns a stable, generalizable rule rather than overfitting train-pool noise. 

\section{Results}
\label{sec:results}

We present results in three parts. We first show that construction choices produce substantial differences in downstream performance, establishing that the construction space is worth searching. Second, we show that decision trees trained on small representative subsets predict seperately drawn construction quality consistently for GNN node classification and GraphRAG retrieval, validating that quality is learnable from subsampled evaluations. Third, we show that reliable construction rankings emerge from small evaluation budgets for both pipelines, with node-classification rankings stable at the smallest budgets and retrieval rankings sharpening as the question budget grows.

\subsection{Construction Performance}
\label{sec:results_construction}

\begin{figure}[!htbp]
\centering
\includegraphics[width=0.95\columnwidth]{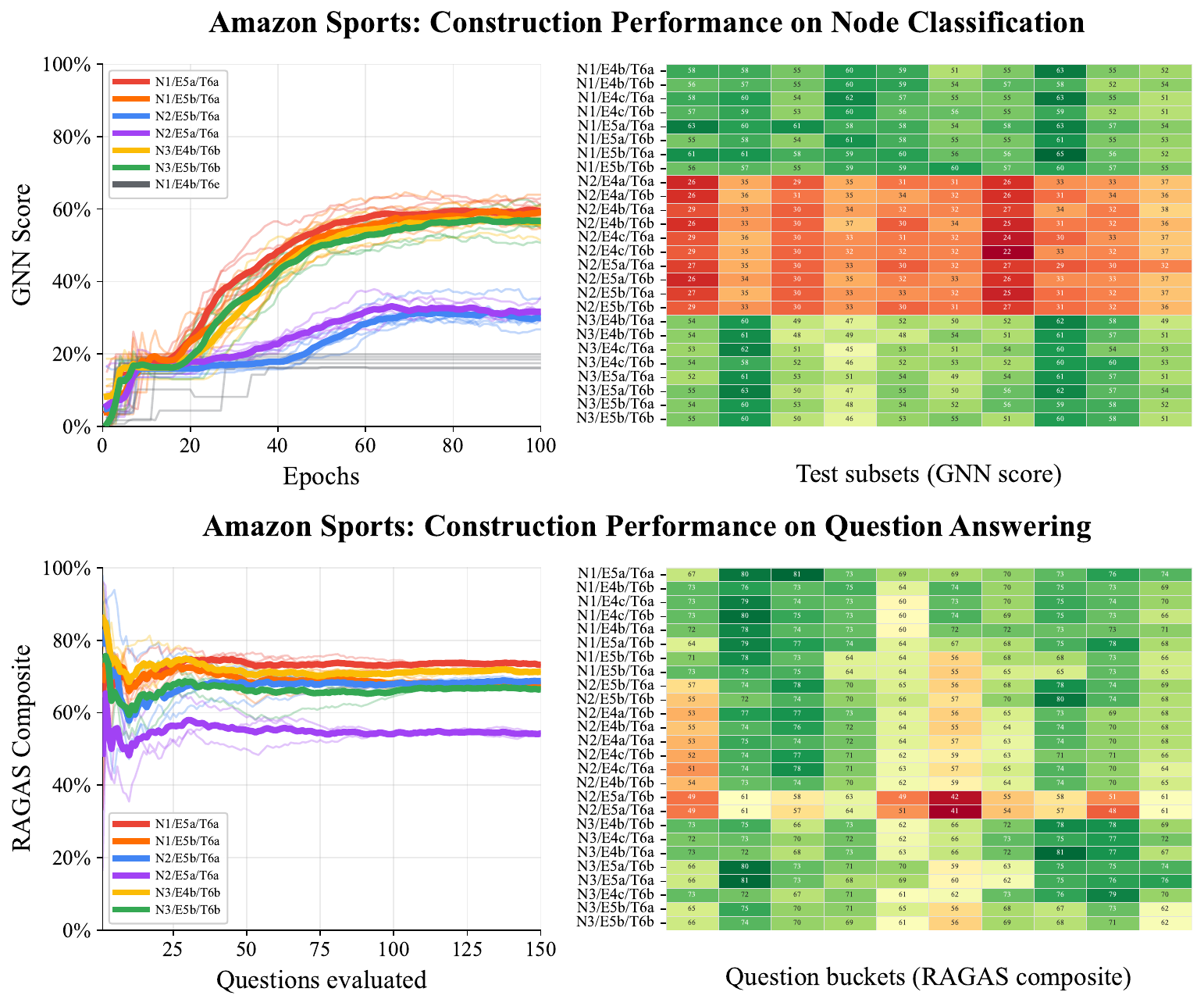}
\caption{Construction performance results for Sports. We visualize 6 of 26 variants, revealing a clear performance split between N2 and N1/N3 for the GNN task (top row). The bottom row has less extreme scoring differences, but similar superiority among variants. The thick lines represent means; thin lines are subsets (top) and question orderings (bottom). On the right, heatmaps show the final scores for each variant and subsample or question bucket. T6e represents the no text control.}
\label{fig:heatmap_amazon}
\end{figure}
\begin{figure}[htbp]
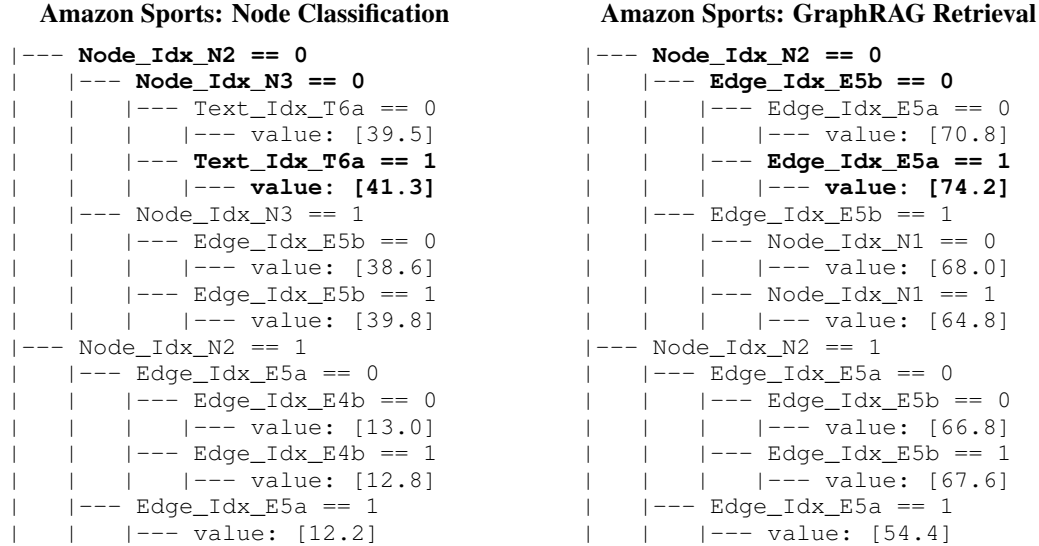

\centering
\begin{minipage}[t]{0.45\textwidth}
\centering
\textbf{Amazon Sports: Node Classification}\\[2pt]
\footnotesize
\begin{alltt}
|--- \textbf{Node_Idx_N2 == 0}
|   |--- \textbf{Node_Idx_N3 == 0}
|   |   |--- Text_Idx_T6a == 0
|   |   |   |--- value: [39.5]
|   |   |--- \textbf{Text_Idx_T6a == 1}
|   |   |   |--- \textbf{value: [41.3]}
|   |--- Node_Idx_N3 == 1
|   |   |--- Edge_Idx_E5b == 0
|   |   |   |--- value: [38.6]
|   |   |--- Edge_Idx_E5b == 1
|   |   |   |--- value: [39.8]
|--- Node_Idx_N2 == 1
|   |--- Edge_Idx_E5a == 0
|   |   |--- Edge_Idx_E4b == 0
|   |   |   |--- value: [13.0]
|   |   |--- Edge_Idx_E4b == 1
|   |   |   |--- value: [12.8]
|   |--- Edge_Idx_E5a == 1
|   |   |--- value: [12.2]
\end{alltt}
\end{minipage}
\hfill
\begin{minipage}[t]{0.45\textwidth}
\centering
\textbf{Amazon Sports: GraphRAG Retrieval}\\[2pt]
\footnotesize
\begin{alltt}
|--- \textbf{Node_Idx_N2 == 0}
|   |--- \textbf{Edge_Idx_E5b == 0}
|   |   |--- Edge_Idx_E5a == 0
|   |   |   |--- value: [70.8]
|   |   |--- \textbf{Edge_Idx_E5a == 1}
|   |   |   |--- \textbf{value: [74.2]}
|   |--- Edge_Idx_E5b == 1
|   |   |--- Node_Idx_N1 == 0
|   |   |   |--- value: [68.0]
|   |   |--- Node_Idx_N1 == 1
|   |   |   |--- value: [64.8]
|--- Node_Idx_N2 == 1
|   |--- Edge_Idx_E5a == 0
|   |   |--- Edge_Idx_E5b == 0
|   |   |   |--- value: [66.8]
|   |   |--- Edge_Idx_E5b == 1
|   |   |   |--- value: [67.6]
|   |--- Edge_Idx_E5a == 1
|   |   |--- value: [54.4]
\end{alltt}
\end{minipage}
\caption{Decision trees fit on the construction performance data with max-depth 3. The bold axes highlight graph construction decisions that lead to the best performing variants. We find that the most meaningful splits are not choosing N2 for both tasks at the top level, but the trees disagree below that split. Both decision trees achieve $R^2>0.99$ on their respective training datasets. }
\label{fig:trees-amazon-gnn-graphrag}
\end{figure}

Downstream node classification performance exhibits structured variation driven by TAG construction choices. As illustrated in Figure~\ref{fig:heatmap_amazon} (left), training curves for Amazon Sports variants show that primary-entity constructions (N1, reviews) and aggregate-entity constructions (N3, product families) converge over 50\% at 100 epochs for node classification, while secondary-entity constructions (N2, reviewers) plateau near 30\%. Reviewer-centric graphs are not the correct construction choice for either task. This is expected, since most Amazon reviewers in the dataset appear exactly once and only a handful act as genuine hubs for many reviews. However, while the same ordering persists in GraphRAG retrieval below, the gap narrows considerably for the new task. N1 and N3 remain statistically indistinguishable, while N2/E5a is noticeably worse. In heatmaps for Figure~\ref{fig:heatmap_amazon} (right), we see the final variant performance differences subsist across independent samples in both tasks, confirming that the gap reflects construction choice rather than sampling variation. Appendix~\ref{app:datasets} reports the same two views for the remaining four datasets and shows that the winning constructions shift with the dataset as well as the task. We exclude T6e from both heatmaps as its thirteen variants all sit at the majority-class floor, as shown on the left for node classification and because question-answering cannot be run without text.

\subsection{Decision Tree Meta-Learner}
\label{sec:results_dt}

The decision tree in Figure ~\ref{fig:trees-amazon-gnn-graphrag} narrows down the components that result in a high performing graph to their specific construction choices. By following the bolded decision axes, we find the best performing graph construction achieves a score of 41.3 by selecting review nodes N1 (N2=0, N3=0) and full contextual text T6a. 
The full corresponding variant here is N1/E5a/T6a, though there is little variance among edge choices for the top performing N1/T6a variants. In this case, one might choose whichever N1/T6a edge variant was the cheapest to compute for the full scale run. On the right side, we see the decision tree for retrieval has diverged below the shared root choice to avoid N2. Instead of prioritizing text fidelity, retrieval branches on edge semantics, selecting E5a to maximize performance. This suggests that while GNN node classification depends heavily on rich textual context, GraphRAG retrieval is driven primarily by relational edge structure here. Crucially, despite high $R^2$, trees fit on training data scores do not inherently generalize to unseen data.

\subsection{Decision Tree Validation} 
\label{sec:results_graphrag}

\begin{table}[htbp]
\centering
\renewcommand{\arraystretch}{1.25}
\small
\setlength{\tabcolsep}{3.7pt}
\caption{Decision Tree Structural Consistency Checks }
\label{tab:dt_consistency}
\vspace{0.5em}

\textbf{GNN-Based Decision Tree Consistency ($^{*}p<0.05$)}

\begin{tabular}[t]{lccccc}
\toprule
$\rho$ & Hist & Sports & ArXiv & Elec & Toys \\
\midrule
Hist  & \textbf{+0.97*} & +0.72* & +0.75* & +0.65* & +0.68* \\
Sports& +0.86* & \textbf{+0.98*} & +0.48\phantom{*} & +0.92* & +0.92* \\
ArXiv & +0.86* & +0.50* & \textbf{+0.97*} & +0.48\phantom{*} & +0.48\phantom{*} \\
Elec  & +0.48\phantom{*} & +0.95* & +0.50\phantom{*} & \textbf{+0.98*} & +0.94* \\
Toys  & +0.65\phantom{*} & +0.93* & +0.47\phantom{*} & +0.93* & \textbf{+0.95*} \\
\bottomrule
\end{tabular}%
\hspace{2em}
\begin{tabular}[t]{lrcc}
\toprule
Dataset & $n$ & Within & Cross \\
\midrule
History                 & 12 & \textbf{0.00} & 0.58 \\
Sports                  & 39 & \textbf{0.05} & 0.46 \\
ArXiv                   & 24 & \textbf{0.09} & 0.43 \\
Electronics\phantom{xx} & 39 & \textbf{0.14} & 0.43 \\
Toys                    & 39 & \textbf{0.28} & 0.41 \\
\bottomrule
\end{tabular}

\vspace{1em}
\textbf{GraphRAG-Based Decision Tree Consistency ($^{*}p<0.05$)}

\begin{tabular}[t]{lccccc}
\toprule
$\rho$ & Hist & Sports & ArXiv & Elec & Toys \\
\midrule
Hist  & \textbf{+0.98*} & +0.03\phantom{*} & -0.12\phantom{*} & +0.04\phantom{*} & +0.18\phantom{*} \\
Sports& +0.79\phantom{*} & \textbf{+0.81*} & +0.67\phantom{*} & +0.73* & +0.25\phantom{*} \\
ArXiv & +0.59\phantom{*} & +0.57\phantom{*} & \textbf{+0.84*} & +0.72* & +0.06\phantom{*} \\
Elec  & +0.57\phantom{*} & +0.76* & +0.81* & \textbf{+0.81*} & +0.13\phantom{*} \\
Toys  & +0.78\phantom{*} & -0.03\phantom{*} & -0.26\phantom{*} & +0.19\phantom{*} & \textbf{+0.94*} \\
\bottomrule
\end{tabular}%
\hspace{2em}
\begin{tabular}[t]{lrcc}
\toprule
Dataset & $n$ & Within & Cross \\
\midrule
History                 & 8  & \textbf{0.21} & 0.59 \\
Sports                  & 26 & \textbf{0.26} & 0.34 \\
ArXiv                   & 16 & \textbf{0.13} & 0.47 \\
Electronics\phantom{xx} & 26 & \textbf{0.23} & 0.34 \\
Toys                    & 26 & \textbf{0.08} & 0.36 \\
\bottomrule
\end{tabular}
\end{table}

We next confirm that our trees reflect stable, generalizable principles in Table~\ref{tab:dt_consistency}. We utilize Spearman $\rho$ and normalized tree edit distance on separately drawn data to verify that learned decision trees consistently predict construction rankings within and across datasets. For the Spearman $\rho$ matrices (left), rows represent training datasets and columns represent target datasets; thus, diagonals measure within-dataset consistency while off-diagonals reflect cross-dataset transfer. Within-dataset consistency is high across both tasks, with every diagonal entry exceeding $0.80$ ($0.90+$ for GNNs) and achieving statistical significance ($^{*}p<0.05$). To confirm that variant rankings are not an artifact of shared support between the training and evaluation subsets, we repeat the analysis on 15 subsets per pool drawn to be strictly disjoint.  Spearman $\rho$ remains within $0.07$ of the overlapping baseline on four of five datasets and improves on History. Cross-dataset transfer is also substantial for node classification, largely because all five datasets agree that text-bearing constructions outperform the no-text control ($T6e$), whereas GraphRAG exhibits high transfer primarily among Amazon datasets with shared schemas. For the tree edit distance evaluations (right), \textit{Within} refers to the normalized edit distance between trees fitted on independent splits of the same dataset, and \textit{Cross} refers to the mean edit distance across all cross-dataset tree pairs. Although independent splits carry no structural guarantees, within-dataset edit distances remain consistently low (averaging 0.11 for GNNs vs. 0.46 for cross-dataset pairs), confirming that decision trees recover highly stable, repeatable rules.

\begin{figure}[htbp]
\centering
\includegraphics[width=\columnwidth]{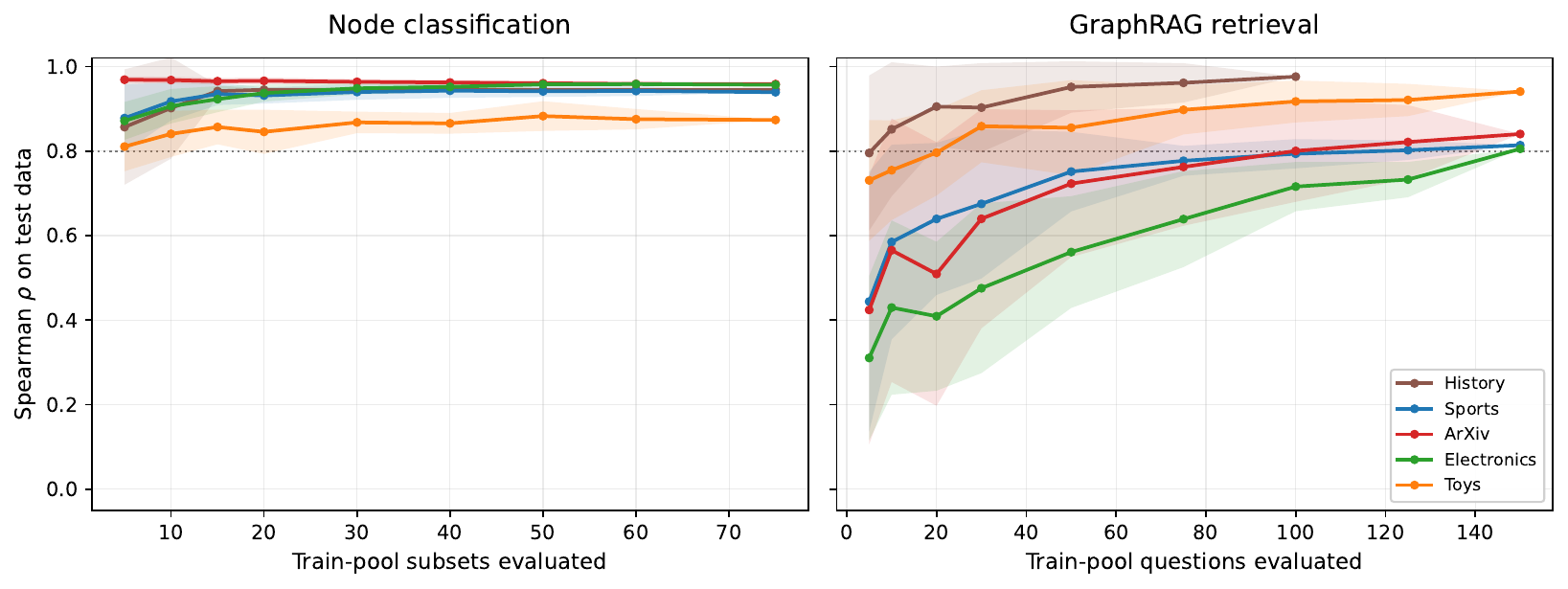}
\caption{Construction ranking stability versus 
evaluation budget for GNN and GraphRAG. 
Left: GNN Spearman $\rho$ against $n$
train-pool subsets across five datasets. 
Right: GraphRAG stability against 
question counts with 20 random seeds. 
Both pipelines show increasing ranking stability with larger evaluation budgets, with GNN starting high and GraphRAG climbing gradually.}

\label{fig:stability_dual}
\end{figure}

TACO-Graph maps the entire valid construction space at a computational cost comparable to training just a few full-scale models. For node classification, early ranking convergence (Figure~\ref{fig:stability_dual}) enables dramatic compute savings: evaluating a small handful of data subsets yields stable decision tree rankings. Consequently, surveying the full space of construction variants across our datasets costs only a small multiple of a single full-scale evaluation, as seen in Table \ref{tab:timing}.

\begin{table*}[h]
\centering
\renewcommand{\arraystretch}{1.25}
\caption{Construction space evaluation cost. 
}
\vspace{5pt}
\label{tab:timing}
\setlength{\tabcolsep}{3.6pt}
\small
\begin{tabular}{@{}l cc cc cc cc@{}}
\toprule
 & \multicolumn{2}{c}{Variants}
 & \multicolumn{2}{c}{Total scoring (hr)}
 & \multicolumn{2}{c}{Full-scale run (hr)}
 & \multicolumn{2}{c}{\textbf{Whole-space cost}} \\
\cmidrule(lr){2-3} \cmidrule(lr){4-5} \cmidrule(lr){6-7} \cmidrule(lr){8-9}
Dataset & GNN & RAG & GNN & RAG & GNN & RAG & GNN & RAG \\
\midrule
ArXiv       & 24 & 16 & 0.14 & 2.8 & 0.05 & 0.11 & $\mathbf{2.7\times}$ & $\mathbf{26.4\times}$ \\
Sports      & 39 & 26 & 0.16 & 3.4 & 0.05 & 0.12 & $\mathbf{3.0\times}$ & $\mathbf{27.5\times}$ \\
History     & 12 &  8 & 0.04 & 0.8 & 0.02 & 0.10 & $\mathbf{2.5\times}$ & $\mathbf{8.0\times}$  \\
Electronics & 39 & 26 & 0.16 & 2.8 & 0.02 & 0.11 & $\mathbf{6.9\times}$ & $\mathbf{24.5\times}$ \\
Toys        & 39 & 26 & 0.16 & 2.8 & 0.02 & 0.10 & $\mathbf{9.3\times}$ & $\mathbf{26.8\times}$ \\
\bottomrule
\end{tabular}
\end{table*}

Table~\ref{tab:timing} expresses total proxy evaluation costs in terms of equivalent full-scale training runs. For Sports GNN node classification, evaluating all 39 variants on proxy subsets takes 0.16 hours, compared to 0.054 hours for a single run on all training data, a total cost of $3.0\times$ full-scale runs. Across datasets, mapping the entire GNN search space costs between $2.5\times$ and $9.3\times$ full-scale runs; the compute budget of one full-scale run thus yields proxy rankings for roughly 4 to 13 variants. In practice, a practitioner testing 3 to 9 full-scale constructions spends the budget required to evaluate the entire search space via our method. Conversely, GraphRAG retrieval operates in a higher cost regime: rankings do not plateau early, so evaluation requires the full 300-question budget; elevating whole-space costs to between $8.0\times$ and $27.5\times$ full-scale runs. For GraphRAG, TACO-Graph only provides structured coverage over the whole design space rather than significant runtime savings.

\section{Discussion and Conclusion}
\label{sec:conclusion}

Construction choice is consequential: across five datasets, alternative TAG constructions yield substantially different GNN classification and GraphRAG performance, aligning with prior findings on feature and schema design~\citep{tape2023, rose2024, autograble2026, rdb2gbench2025}. The construction space is learnable: decision trees trained on small representative samples predict construction quality on separately drawn data  with mean Spearman $\rho > 0.9$ for node classification and $\rho > 0.8$ for GraphRAG retrieval across all five datasets. This extends surrogate-model search to graph construction, offering a predictive alternative to agent-driven methods~\citep{hutter2019automl, autog2025}. Rather than relying on a fixed universal schema~\citep{schemagraphs2026}, construction rules are best learned empirically and per dataset, as within-dataset consistency strictly exceeds cross-dataset transfer. Crucially, these rankings emerge from minimal evaluation budgets. Sampling just a few subsets ($<5\%$ of each full dataset) and 300 questions allows TACO-Graph to survey the entire construction space at the cost of a few full-scale evaluations on an arbitrary variant.

In addition to finding rules for each dataset and showing that they generalize, we also catalog what hinders learning tree construction rules. The History dataset, which contains only N1 node variants, has the smallest construction space in our study and the flattest response for node classification: its eight variants span 6.3 scoring points. With so little to separate the best construction from the worst, systematic search buys a practitioner almost nothing on that task, and TACO-Graph reports this before any full-scale run. However, History's retrieval scores span 22.0 points, so construction choice remains crucial for GraphRAG (Appendix \ref{app:datasets}). ArXiv inverts this pattern (46 point classification range vs. 9 point retrieval range), proving that search utility is bounded by the task-dataset pairing rather than the dataset alone. More broadly, decision trees require a meaningful score spread across variants to learn useful rules, regardless of whether the performance level is on the order of Toys ($0.01$ to $0.12$ raw GNN score) or ArXiv ($0.38$ to $0.84$). For GraphRAG, although individual RAGAS axes vary in discriminative power (context relevance highest, answer relevance lowest), composite aggregation provides the most robust ranking signal; accelerating this evaluation via lighter-weight scoring proxies is a direction for future work.

In conclusion, TACO-Graph provides a systematic, interpretable, and transferable framework for enumerating, scoring, and selecting text-attributed graph constructions for downstream tasks. Future work includes extending the evaluation to additional GNN task types and non-tabular text corpora, investigating zero-cost graph structural proxies as cheaper alternatives to GNN scoring for construction selection, and applying the framework to multi-table relational databases where the construction space is richer. The edge axis can also be widened by decomposing single edge types into distinct semantic relations \citep{rose2024}. Knowledge graphs are another natural target since our scoring and tree fitting do not depend on how the graph was obtained, though the construction space would range over extraction choices rather than schema columns. Taken together, these directions position TACO-Graph as a framework for principled construction search across increasingly diverse graph-structured data. We believe that with larger and more diverse datasets, TACO-Graph can achieve consistent decision tree quality across a broader array of downstream tasks, providing an automated, interpretable rationale for design choices in graph corpora.

\section*{Reproducibility Statement}
All code, dataset configuration files, and trained decision tree artifacts are
available at \url{https://github.com/AmiriHayes/text_attributed_graphs}.
Section 3 describes the sampling procedure and scoring formula in full.
Appendix A provides dataset-specific notes, including label design and
question generation decisions.

\subsection*{AI Use Statement}
We used generative AI to generate synthetic data sets: the evaluation
questions described in Section 3.2.2 were generated with GPT-4o-mini,
which also served as the generator and judge for RAGAS scoring. We used
generative AI (Claude, Anthropic; ChatGPT, OpenAI) for code generation,
literature search, and writing assistance throughout this project. All
code was verified by the authors, and all empirical claims were checked.
The authors take full responsibility for this work.

\bibliography{main}
\bibliographystyle{iclr2027_conference}

\clearpage
\appendix

\section{Dataset-Specific Notes}
\label{app:datasets}

\begin{figure}[!t]
\centering

\begin{minipage}{0.49\textwidth}
  \centering
  \footnotesize\textbf{ArXiv GNN Construction Performance}\\[0.2em]
  \includegraphics[width=\linewidth]{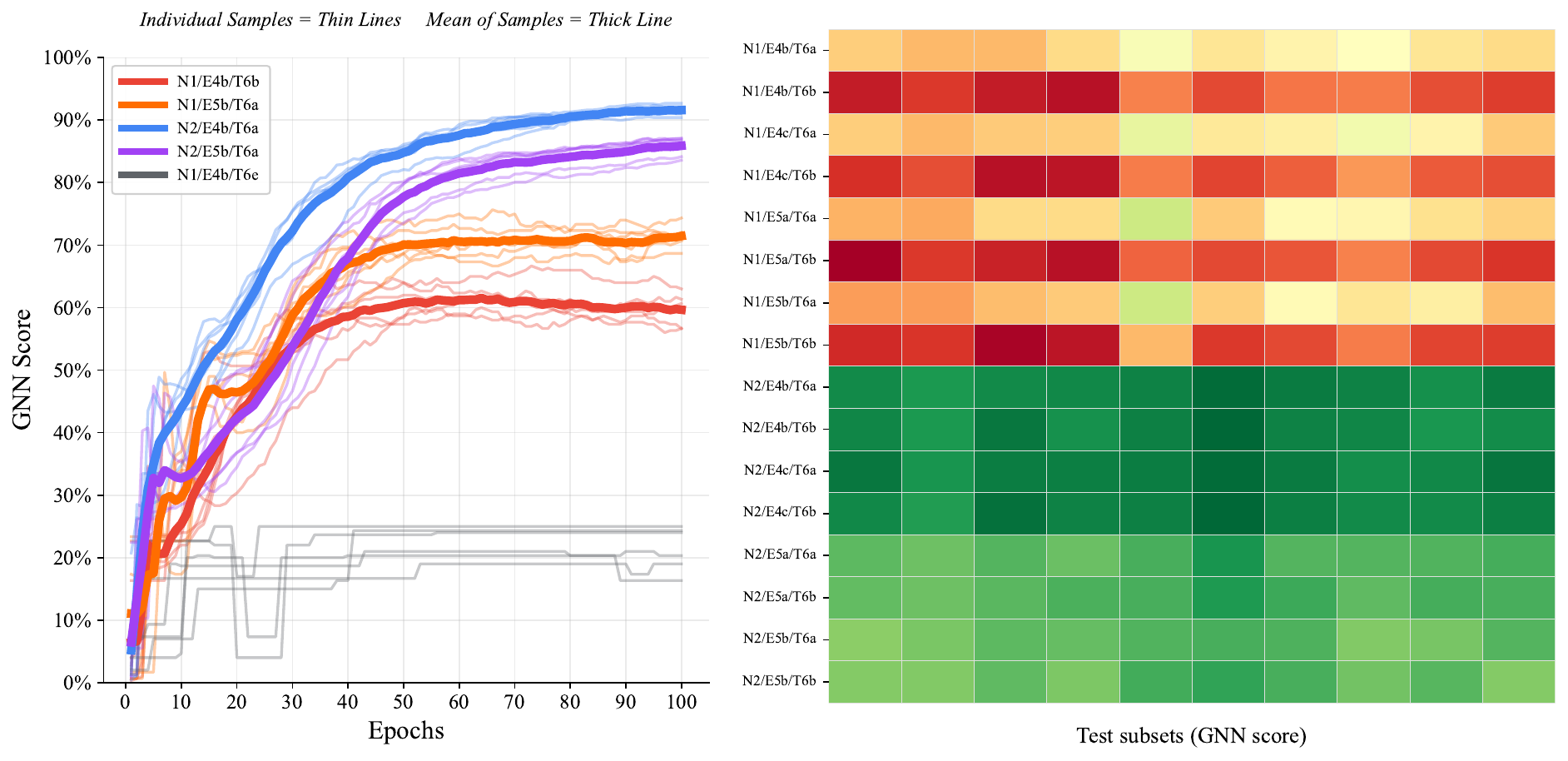}
\end{minipage}%
\hfill
\begin{minipage}{0.49\textwidth}
  \centering
  \footnotesize\textbf{ArXiv GraphRAG Construction Performance}\\[0.2em]
  \includegraphics[width=\linewidth]{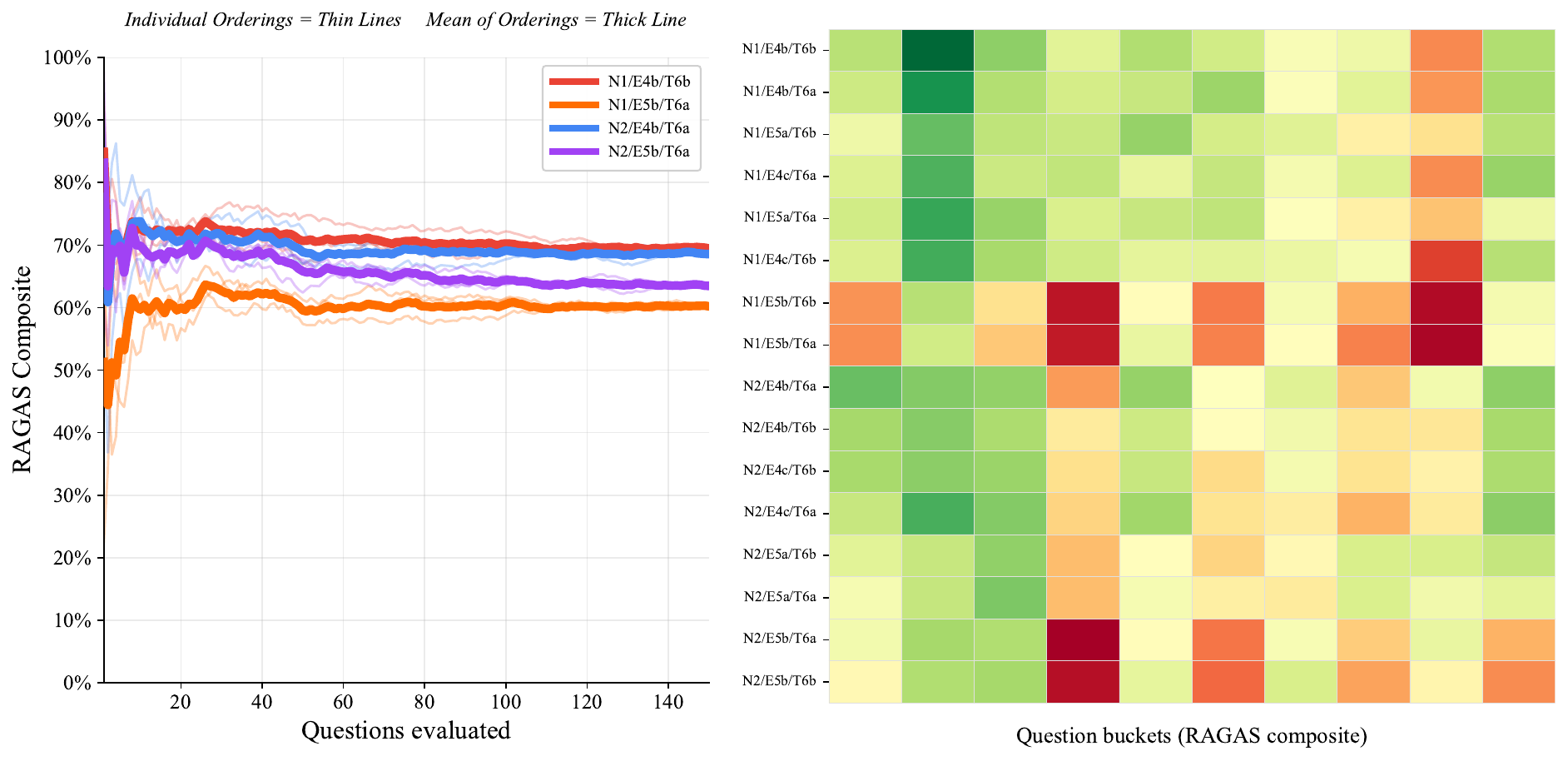}
\end{minipage}

\vspace{0.3em}

\begin{minipage}{0.49\textwidth}
  \centering
  \footnotesize\textbf{History GNN Construction Performance}\\[0.2em]
  \includegraphics[width=\linewidth]{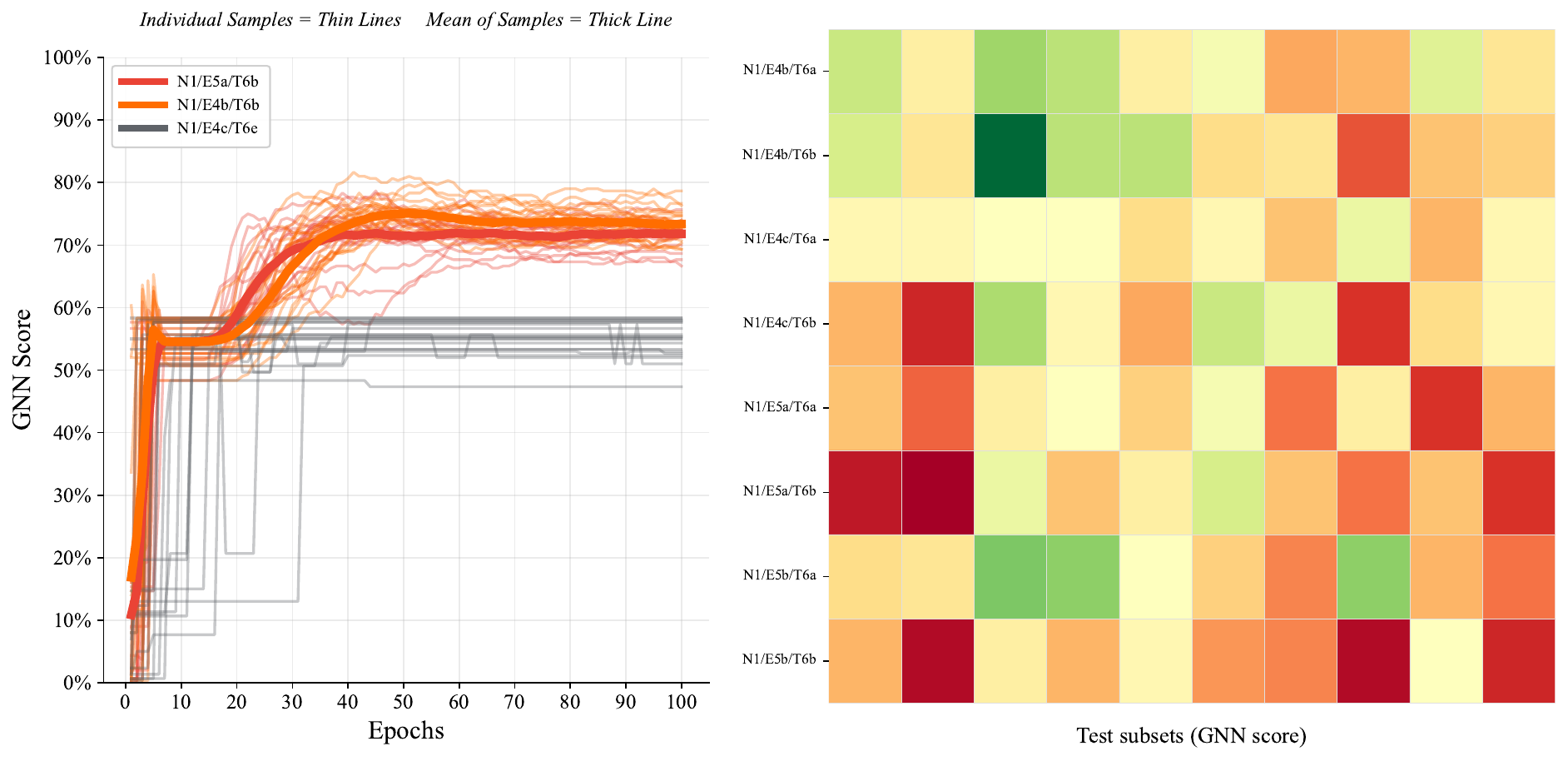}
\end{minipage}%
\hfill
\begin{minipage}{0.49\textwidth}
  \centering
  \footnotesize\textbf{History GraphRAG Construction Performance}\\[0.2em]
  \includegraphics[width=\linewidth]{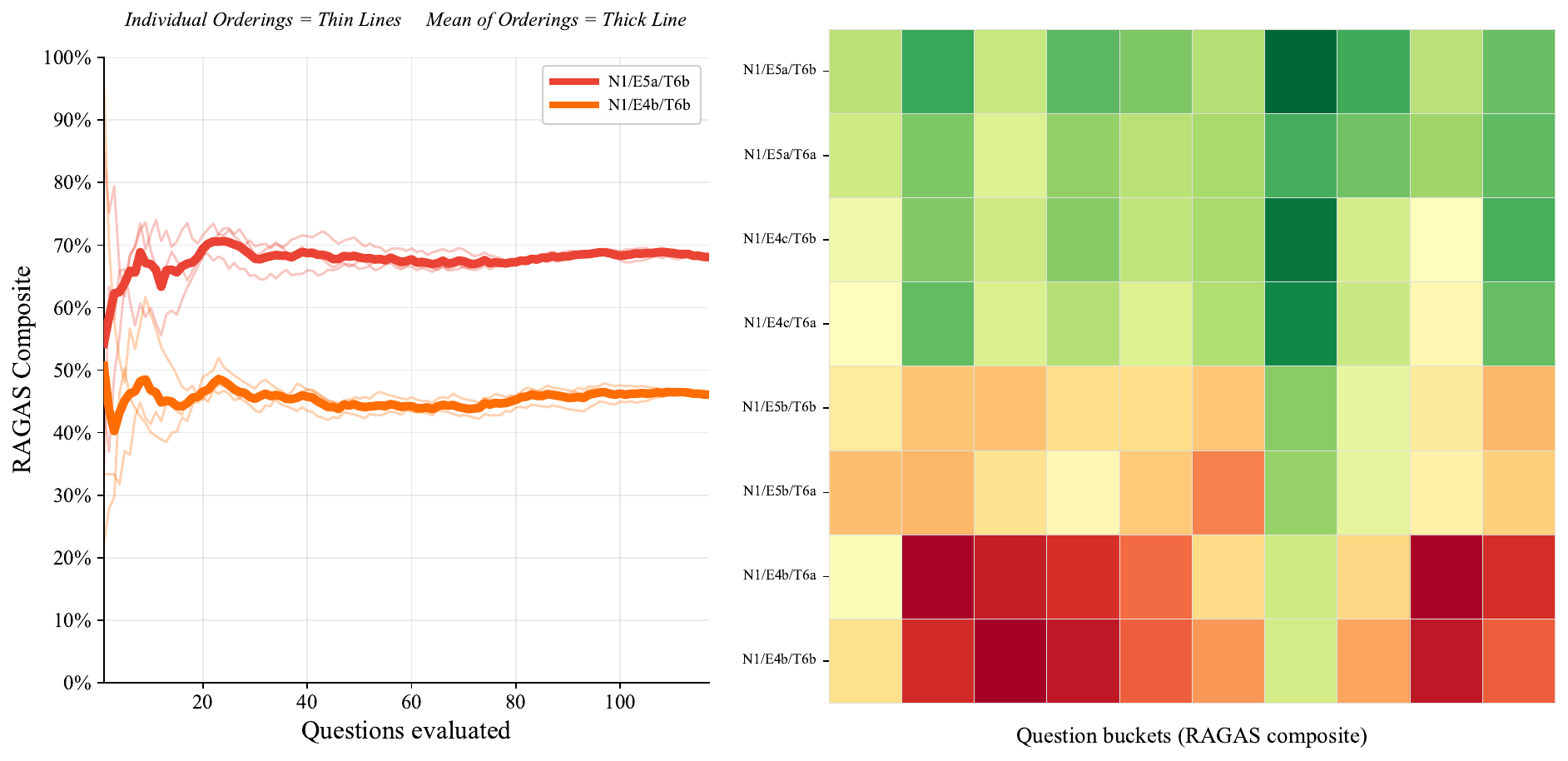}
\end{minipage}

\vspace{0.3em}

\begin{minipage}{0.49\textwidth}
  \centering
  \footnotesize\textbf{Electronics GNN Construction Performance}\\[0.2em]
  \includegraphics[width=\linewidth]{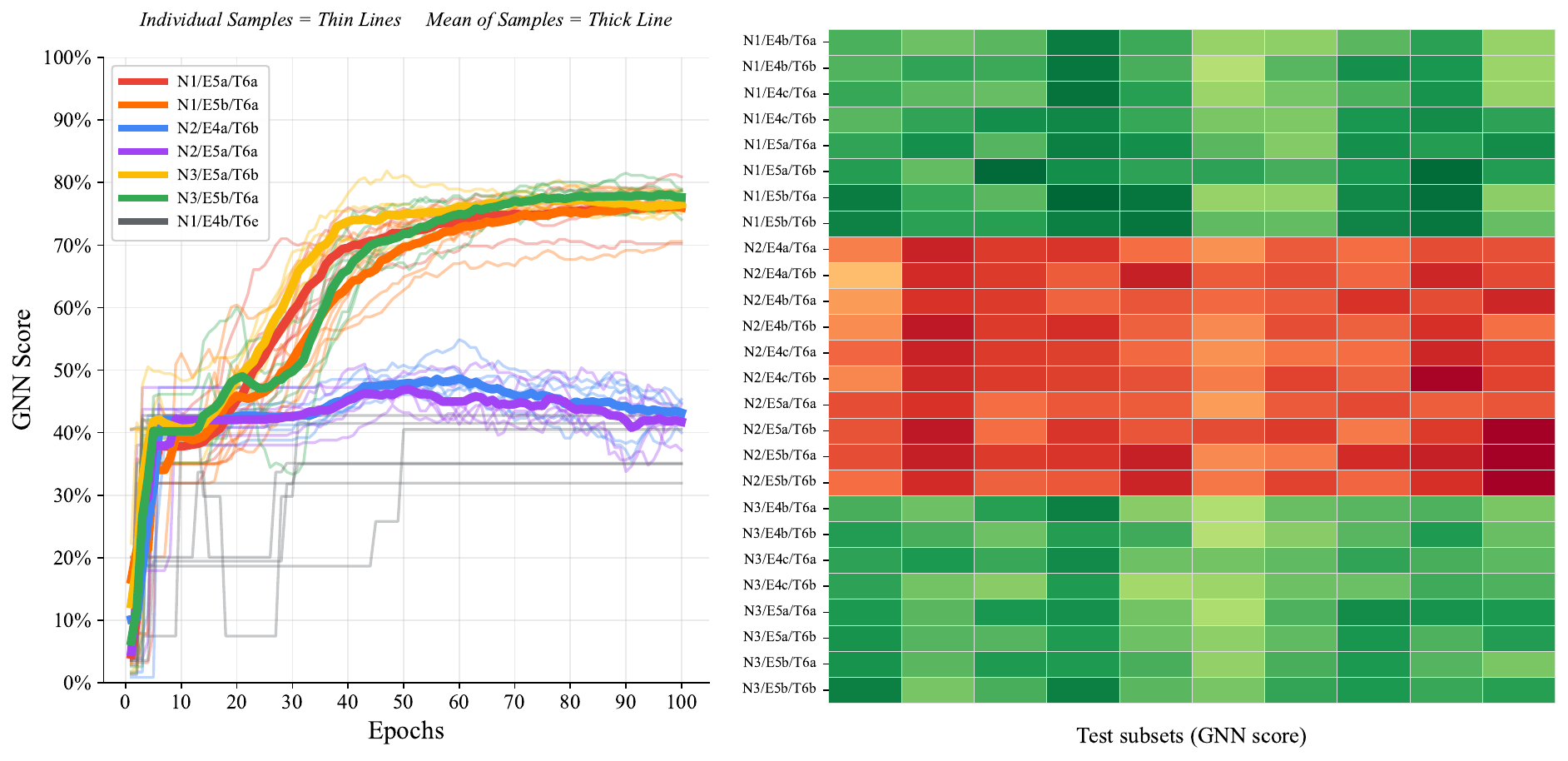}
\end{minipage}%
\hfill
\begin{minipage}{0.49\textwidth}
  \centering
  \footnotesize\textbf{Electronics GraphRAG Construction Performance}\\[0.2em]
  \includegraphics[width=\linewidth]{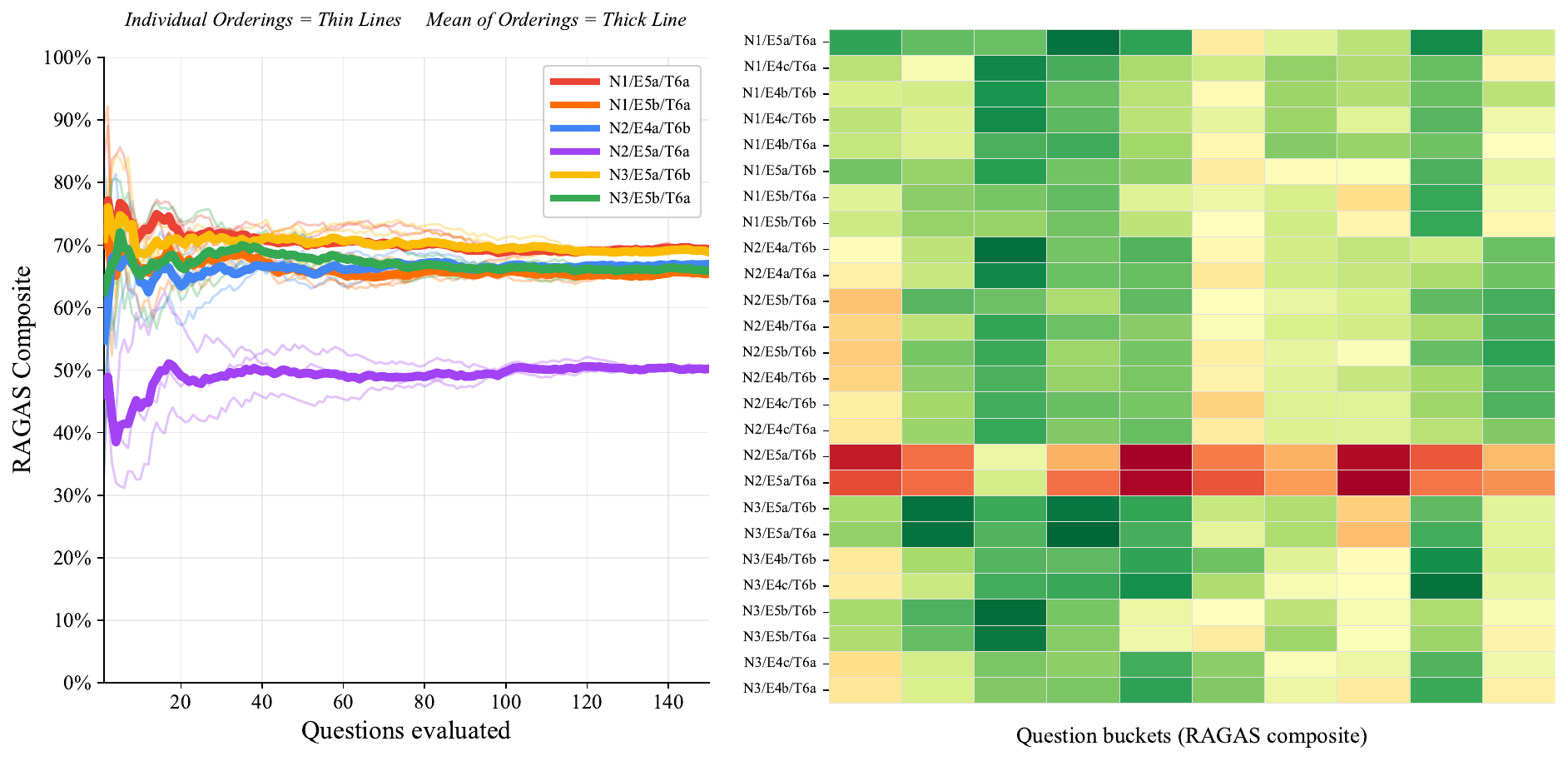}
\end{minipage}

\vspace{0.3em}

\begin{minipage}{0.49\textwidth}
  \centering
  \footnotesize\textbf{Toys GNN Construction Performance}\\[0.2em]
  \includegraphics[width=\linewidth]{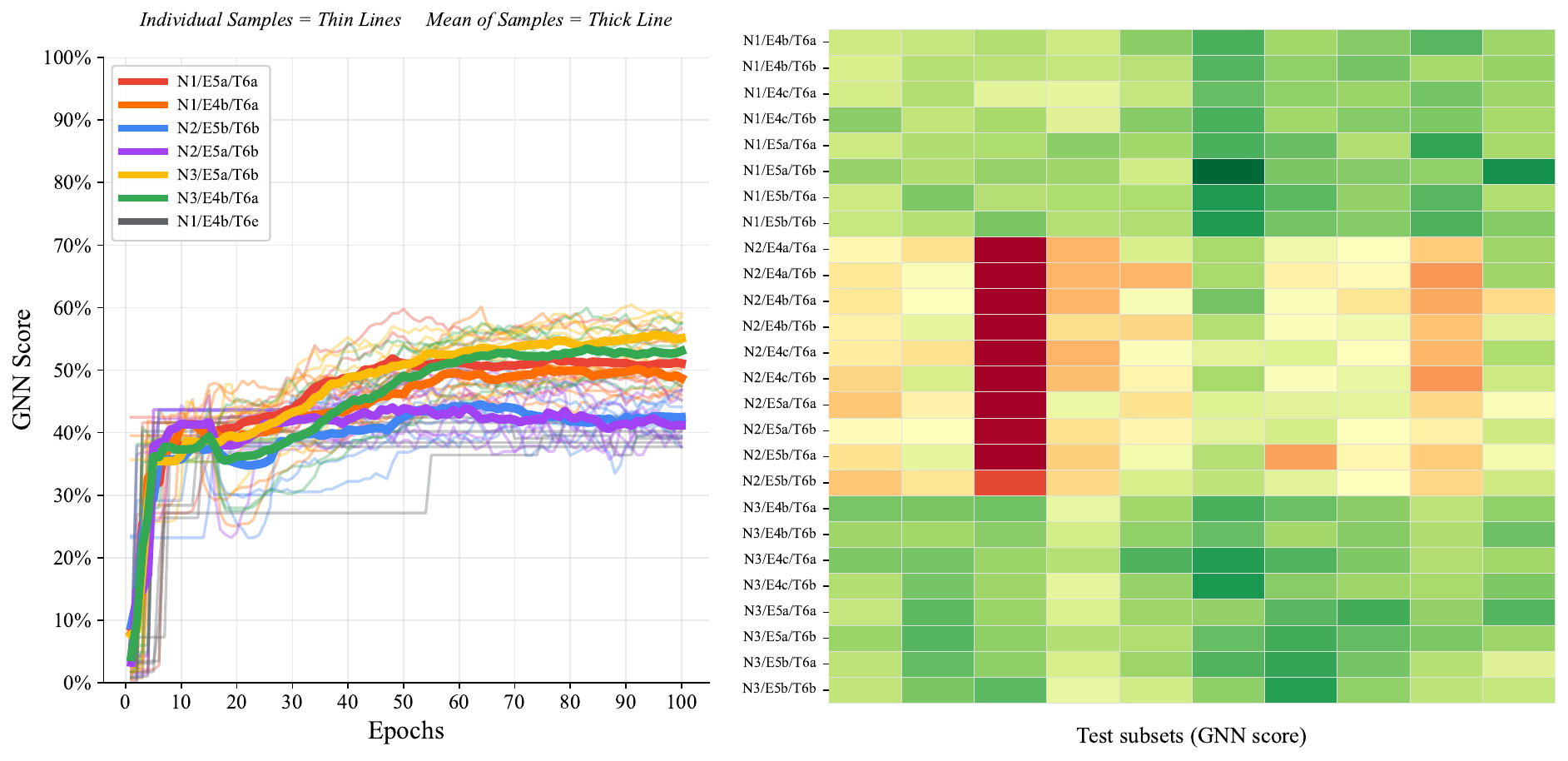}
\end{minipage}%
\hfill
\begin{minipage}{0.49\textwidth}
  \centering
  \footnotesize\textbf{Toys GraphRAG Construction Performance}\\[0.2em]
  \includegraphics[width=\linewidth]{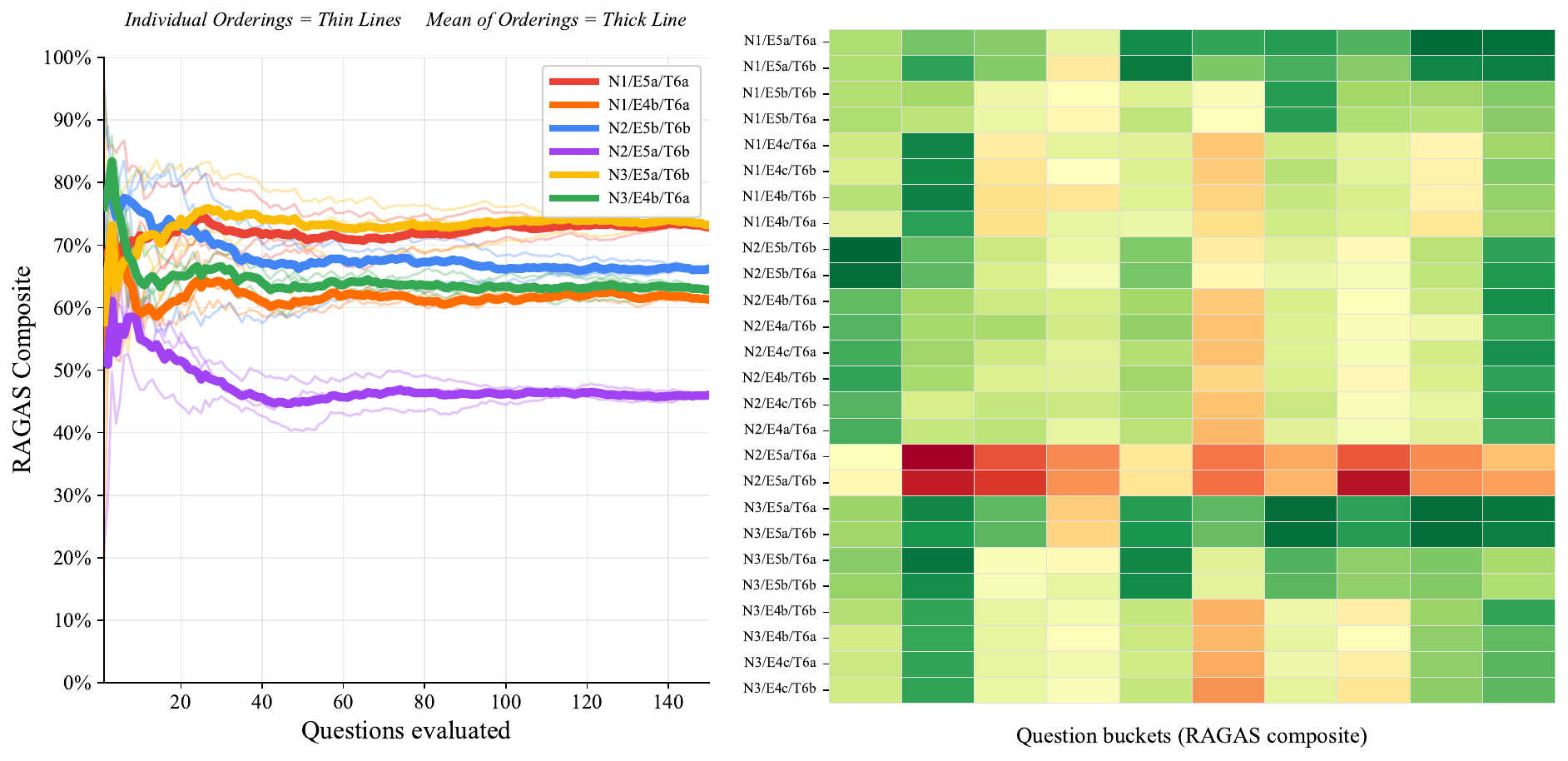}
\end{minipage}

\caption{Construction performance for the four datasets not shown in the main
body. Rows are datasets; the left column is node classification and the right
column is GraphRAG retrieval. Within each panel, scores accumulate over
evaluation on the left and final per-variant scores appear as a heatmap on the
right.}
\label{fig:construction_four_datasets}
\end{figure}

Figure~\ref{fig:construction_four_datasets} plots construction performance for
the four datasets not shown in the main body, and the paragraphs below describe
each dataset individually. Two conventions apply throughout. Consistency
figures quoted per dataset exclude the no-text control, so they sit slightly
below the corresponding diagonal entries of Table~\ref{tab:dt_consistency},
which includes it. Variant counts quoted per dataset likewise exclude the
control and the empirically degenerate variants removed by the filter of
Section~\ref{sec:ontology}.

\paragraph{\textbf{Construction parameterizations.}}
\label{app:params}
Node text is embedded with SBERT \texttt{all-MiniLM-L6-v2} into 384 dimensions.
T6a concatenates the primary content with its surrounding context, T6b uses the
primary content alone, and T6e replaces the embedding with an all-zero vector,
so the control is structure-only rather than text-free but still featured. E4a
links nodes sharing an aggregate identifier; E4b and E4c link nodes through a
shared participation identifier, weighted and unweighted respectively; E5a
links pairs whose text embeddings reach cosine similarity $0.75$; E5b links
each node to the five nodes closest to it in degree centrality. A variant whose
graph exceeds $200{,}000$ undirected edges is recorded as degenerate and not
trained, which is the operational form of pruning constraint (2). Node
classification is scored by a floored McFadden pseudo-$R^2$,
$S = \max(0,\, 1 - \mathrm{LL}_{\text{model}}/\mathrm{LL}_{\text{null}})$, with
an intercept-only null fit on the empirical class frequencies of that subset's
test mask; variants below the null baseline are clamped at zero.

\paragraph{\textbf{GraphRAG construction and question generation.}}
Each variant is indexed with a LlamaIndex \texttt{VectorStoreIndex} over
\texttt{text-embedding-3-small}. A query retrieves the top three passages by
dense similarity, which are then expanded one hop along that variant's edges
and capped at fifteen passages. Questions are generated per dataset in three
subtypes of one hundred. Validation is a per-subtype pilot scored against a
flat vector index that is independent of any graph construction, so the
benchmark is not biased toward a particular variant; a subtype whose pilot
falls below $0.5$ faithfulness is dropped in full, which is why History retains
$233$ questions ($100/100/33$) rather than losing individual items.

\paragraph{\textbf{ArXiv (Citation Network).}}
ArXiv is a citation network of preprints spanning physics, mathematics,
computer science, quantitative biology, quantitative finance, and statistics,
where each node carries a title and abstract. Sixteen variants survive across
two node types, four edge types, and two text fidelities; N3 is excluded due to
label overlap and T6e is a baseline. Node classification predicts top-level
subject category across 18 classes, which the GNN learned well in only 100 epochs (scores spanning
38.3 to 84.6). Question generation used shared authorship ($0.78$ faithfulness;
baselines: $0.78$ faithfulness, $0.63$ answer relevance, $0.60$ context
relevance). The classification tree assigned 88\% importance to node type,
while retrieval split first on structural similarity edges (E5b) with 80\%
importance on edge type. Consistency was high ($\rho$ of $0.958$ and $0.840$;
edit distances of $0.045$ and $0.133$ against a cross-dataset mean near
$0.47$).

\paragraph{\textbf{Amazon Sports (E-Commerce).}}
Amazon Sports is a product review corpus covering sporting goods, with nodes
carrying review text and product metadata. Twenty-six variants survive across
three node types, five edge types, and two text fidelities. Characterization
flagged a binary collapse in \texttt{categorical\_label} prior to the level-3
category join. Node classification predicts product subcategory across 47
classes (GNN scores: 9.9 to 42.1). Question generation used shared reviewers
($0.82$ faithfulness; baselines: $0.82$ faithfulness, $0.58$ answer relevance,
$0.65$ context relevance). The classification tree placed all importance on
node type (splitting on N2), while retrieval split importance between edge type
(62\%) and node type (38\%). Consistency was high ($\rho$ of $0.939$ and
$0.814$; edit distances of $0.053$ and $0.261$ against cross-dataset means of
$0.50$ and $0.34$).

\paragraph{\textbf{History (Bibliographic).}}
History is a catalog of history books, where nodes carry titles and
descriptions. Only eight variants survive across one node type, four edge
types, and two text fidelities; lacking author or reviewer columns, N2 and N3
cannot be constructed. Node classification predicts book category across 12
classes, which the GNN learned poorly (scores: 31.0 to 37.3). Question
generation used shared subject classification ($0.72$ faithfulness). The
aggregate subtype failed its validation pilot and was dropped, leaving 233
questions and baselines of $0.72$ faithfulness, $0.57$ answer relevance, and
$0.45$ context relevance, the lowest overall. Uniquely, both trees split first
on edge type. High consistency ($\rho$ of $0.945$ and $0.976$) reflects only
eight variants for this dataset.

\paragraph{\textbf{Electronics (E-Commerce).}}
Electronics is a product review corpus covering consumer technology.
Twenty-six variants survive across three node types, five edge types, and two
text fidelities. Characterization flagged a dominant majority class (81.5\%)
and dense ground-truth edges (74.4\%), making categorical edges near-complete.
Node classification predicts product subcategory across 11 classes (GNN scores:
6.2 to 56.7). Question generation used shared reviewers ($0.79$ faithfulness;
baselines: $0.79$ faithfulness, $0.57$ answer relevance, $0.60$ context
relevance). The classification tree placed all importance on node type
(splitting on N2), while retrieval assigned 77\% to edge type. Consistency was
high ($\rho$ of $0.957$ and $0.805$; edit distances of $0.167$ and $0.229$
against cross-dataset means of $0.46$ and $0.34$).

\paragraph{\textbf{Toys (E-Commerce).}}
Toys is a product review corpus covering toys and games. Twenty-six variants
survive across three node types, five edge types, and two text fidelities.
Characterization flagged extreme class imbalance (86.9\% majority class). Node
classification predicts product subcategory across 16 classes, which the GNN
learned poorly (scores: 1.7 to 12.5). Question generation used shared reviewers
($0.78$ faithfulness; baselines: $0.78$ faithfulness, $0.53$ answer relevance,
$0.58$ context relevance). The classification tree placed 96\% importance on
node type (splitting on N2), while retrieval gave 80\% to edge type.
Consistency was high ($\rho$ of $0.873$ and $0.941$; edit distances of $0.174$
and $0.083$ against cross-dataset means of $0.44$ and $0.36$).

\paragraph{\textbf{Scope of the evaluation.}}
Every score reported here is measured on $1{,}000$-row subsets drawn from
per-dataset pools of 20k to 50k rows, and those pools are themselves samples of
larger corpora. Our claims therefore concern the ordering of constructions
within that regime. Because attribute-matching edge rules grow denser with pool
size while $k$-nearest-neighbour rules do not, absolute score ranges should be
read as properties of the proxy rather than of the dataset, and the magnitude
of a gap between constructions is less portable across scales than its
direction.

\paragraph{\textbf{Portability to a new dataset.}}
In its current state, TACO-Graph divides labour at a single handoff: once a
dataset is expressed in the generic schema and its valid-variant rules are
written, everything downstream runs without further input. The practitioner
supplies three things. First, a column mapping onto the generic keys: the
primary, secondary, and aggregate identifiers, the text fidelities, and the
task label. Second, a label space, since the raw categorical column is
sometimes unusable as given: in Amazon Sports it collapses to two values, and
the 47-class target is recovered by rejoining a category file, so how to bin or
rejoin classes is a modelling choice rather than a mechanical one. Third, the
set of valid variants, which encodes the pruning constraints of
Section~\ref{sec:ontology} for that schema, and in particular which node and
edge pairings would leak the label. From there the pipeline takes over:
characterization reports structural properties, the registry enumerates the
surviving space, and the builder constructs, scores, and tree-fits every
variant automatically.

Two caveats bound this. The automated checks are diagnostic rather than
corrective and do not gate execution: they flag a singleton-heavy secondary
entity, a collapsed or heavily imbalanced label, redundant text fidelities, and
sparse ground-truth edges, but binning classes, transforming a heavy-tailed
scalar, or dropping an unviable node type remains manual. A dataset left
unadjusted will still run and simply produce a flat score spread, leaving the
tree with nothing informative to split on. Second, some checks are previews
rather than guarantees: scalar and semantic edge sparsity cannot be evaluated
before embeddings exist and are reported as unevaluated rather than estimated,
so only a full build clears them. The edge-density guard and the zero-score
filter catch the remaining degenerate cases at scoring time. Automating these three inputs, particularly deriving the validity rules from
the schema rather than hand-writing them, is the natural next step.

\section{Decision Tree Visualizations}
\label{app:trees}

Figure~\ref{fig:trees-appendix} shows one tree per task for each of the four
datasets not covered in the main body, fit on train-pool variant means using
the floored McFadden score for node classification and the RAGAS composite for
retrieval, both on a 0--100 scale. Depth is capped at three for readability
(\texttt{max\_depth=3}, \texttt{min\_samples\_leaf=2}), the no-text control is
excluded because a tree rooted on ``is this the control'' encodes no
construction rule, and the reported $R^2$ is in-sample. These trees therefore
illustrate the learned rules rather than reproduce the depth-eight trees behind
Table~\ref{tab:dt_consistency}.

Electronics and Toys behave like Amazon Sports: both tasks split first on node
type and isolate N2, and the penalty for that choice is severe for
classification but mild for retrieval. ArXiv is the only dataset where the two
tasks disagree at the root, with classification splitting on node type and
retrieval on structural similarity edges. History splits on edge type for both
tasks, but on different edge types, which is consistent with it being the only
dataset with a single node type and therefore no node axis to split on.

\begin{figure*}[t]
\centering
\scriptsize
\begin{minipage}[t]{0.48\textwidth}
\centering\textbf{ArXiv, node classification} ($n{=}16$, $R^2{=}1.000$)
\begin{verbatim}
|--- Node_Idx_N1 == 0
|   |--- Edge_Idx_E5b == 0
|   |   |--- Edge_Idx_E5a == 0
|   |   |   |--- value: [84.09]
|   |   |--- Edge_Idx_E5a == 1
|   |   |   |--- value: [77.03]
|   |--- Edge_Idx_E5b == 1
|   |   |--- value: [74.61]
|--- Node_Idx_N1 == 1
|   |--- Text_Idx_T6a == 0
|   |   |--- value: [38.49]
|   |--- Text_Idx_T6a == 1
|   |   |--- value: [54.15]
\end{verbatim}
\end{minipage}
\hfill
\begin{minipage}[t]{0.48\textwidth}
\centering\textbf{ArXiv, GraphRAG retrieval} ($n{=}16$, $R^2{=}0.990$)
\begin{verbatim}
|--- Edge_Idx_E5b == 0
|   |--- Node_Idx_N2 == 0
|   |   |--- Edge_Idx_E5a == 0
|   |   |   |--- value: [69.95]
|   |   |--- Edge_Idx_E5a == 1
|   |   |   |--- value: [70.68]
|   |--- Node_Idx_N2 == 1
|   |   |--- Edge_Idx_E5a == 0
|   |   |   |--- value: [69.76]
|   |   |--- Edge_Idx_E5a == 1
|   |   |   |--- value: [65.76]
|--- Edge_Idx_E5b == 1
|   |--- Node_Idx_N2 == 0
|   |   |--- value: [59.99]
|   |--- Node_Idx_N2 == 1
|   |   |--- value: [65.40]
\end{verbatim}
\end{minipage}
\vspace{0.4em}
\begin{minipage}[t]{0.48\textwidth}
\centering\textbf{Electronics, node classification} ($n{=}26$, $R^2{=}0.999$)
\begin{verbatim}
|--- Node_Idx_N2 == 0
|   |--- Edge_Idx_E5a == 0
|   |   |--- Edge_Idx_E5b == 0
|   |   |   |--- value: [53.72]
|   |   |--- Edge_Idx_E5b == 1
|   |   |   |--- value: [55.51]
|   |--- Edge_Idx_E5a == 1
|   |   |--- Node_Idx_N3 == 0
|   |   |   |--- value: [56.44]
|   |   |--- Node_Idx_N3 == 1
|   |   |   |--- value: [56.32]
|--- Node_Idx_N2 == 1
|   |--- Edge_Idx_E4a == 0
|   |   |--- Edge_Idx_E5b == 0
|   |   |   |--- value: [12.41]
|   |   |--- Edge_Idx_E5b == 1
|   |   |   |--- value: [11.16]
|   |--- Edge_Idx_E4a == 1
|   |   |--- value: [13.77]
\end{verbatim}
\end{minipage}
\hfill
\begin{minipage}[t]{0.48\textwidth}
\centering\textbf{Electronics, GraphRAG retrieval} ($n{=}26$, $R^2{=}0.972$)
\begin{verbatim}
|--- Node_Idx_N2 == 0
|   |--- Edge_Idx_E5b == 0
|   |   |--- Edge_Idx_E5a == 0
|   |   |   |--- value: [66.20]
|   |   |--- Edge_Idx_E5a == 1
|   |   |   |--- value: [68.29]
|   |--- Edge_Idx_E5b == 1
|   |   |--- Node_Idx_N3 == 0
|   |   |   |--- value: [64.20]
|   |   |--- Node_Idx_N3 == 1
|   |   |   |--- value: [63.53]
|--- Node_Idx_N2 == 1
|   |--- Edge_Idx_E5a == 0
|   |   |--- Edge_Idx_E5b == 0
|   |   |   |--- value: [65.19]
|   |   |--- Edge_Idx_E5b == 1
|   |   |   |--- value: [64.43]
|   |--- Edge_Idx_E5a == 1
|   |   |--- value: [52.06]
\end{verbatim}
\end{minipage}
\vspace{0.4em}
\begin{minipage}[t]{0.48\textwidth}
\centering\textbf{Toys, node classification} ($n{=}26$, $R^2{=}0.995$)
\begin{verbatim}
|--- Node_Idx_N2 == 0
|   |--- Edge_Idx_E5a == 0
|   |   |--- Node_Idx_N1 == 0
|   |   |   |--- value: [8.29]
|   |   |--- Node_Idx_N1 == 1
|   |   |   |--- value: [8.87]
|   |--- Edge_Idx_E5a == 1
|   |   |--- Node_Idx_N3 == 0
|   |   |   |--- value: [10.81]
|   |   |--- Node_Idx_N3 == 1
|   |   |   |--- value: [10.48]
|--- Node_Idx_N2 == 1
|   |--- Edge_Idx_E5a == 0
|   |   |--- Edge_Idx_E5b == 0
|   |   |   |--- value: [1.18]
|   |   |--- Edge_Idx_E5b == 1
|   |   |   |--- value: [1.50]
|   |--- Edge_Idx_E5a == 1
|   |   |--- value: [0.85]
\end{verbatim}
\end{minipage}
\hfill
\begin{minipage}[t]{0.48\textwidth}
\centering\textbf{Toys, GraphRAG retrieval} ($n{=}26$, $R^2{=}0.980$)
\begin{verbatim}
|--- Node_Idx_N2 == 0
|   |--- Edge_Idx_E5a == 0
|   |   |--- Edge_Idx_E5b == 0
|   |   |   |--- value: [58.03]
|   |   |--- Edge_Idx_E5b == 1
|   |   |   |--- value: [68.13]
|   |--- Edge_Idx_E5a == 1
|   |   |--- Node_Idx_N3 == 0
|   |   |   |--- value: [72.48]
|   |   |--- Node_Idx_N3 == 1
|   |   |   |--- value: [73.70]
|--- Node_Idx_N2 == 1
|   |--- Edge_Idx_E5a == 0
|   |   |--- Edge_Idx_E4b == 0
|   |   |   |--- value: [61.34]
|   |   |--- Edge_Idx_E4b == 1
|   |   |   |--- value: [60.82]
|   |--- Edge_Idx_E5a == 1
|   |   |--- value: [45.82]
\end{verbatim}
\end{minipage}
\vspace{0.4em}
\begin{minipage}[t]{0.48\textwidth}
\centering\textbf{History, node classification} ($n{=}8$, $R^2{=}0.926$)
\begin{verbatim}
|--- Edge_Idx_E4b == 0
|   |--- Text_Idx_T6a == 0
|   |   |--- value: [31.65]
|   |--- Text_Idx_T6a == 1
|   |   |--- value: [34.78]
|--- Edge_Idx_E4b == 1
|   |--- value: [39.14]
\end{verbatim}
\end{minipage}
\hfill
\begin{minipage}[t]{0.48\textwidth}
\centering\textbf{History, GraphRAG retrieval} ($n{=}8$, $R^2{=}0.997$)
\begin{verbatim}
|--- Edge_Idx_E5a == 0
|   |--- Edge_Idx_E4c == 0
|   |   |--- Edge_Idx_E5b == 0
|   |   |   |--- value: [47.71]
|   |   |--- Edge_Idx_E5b == 1
|   |   |   |--- value: [51.28]
|   |--- Edge_Idx_E4c == 1
|   |   |--- value: [62.64]
|--- Edge_Idx_E5a == 1
|   |--- value: [68.50]
\end{verbatim}
\end{minipage}
\vspace{0.4em}
\caption{Decision trees for the four datasets not shown in
Figure~\ref{fig:trees-amazon-gnn-graphrag}, fit at the same settings. Left
column is node classification; right column is GraphRAG retrieval; both are
scaled to 0--100.}
\label{fig:trees-appendix}
\end{figure*}

\end{document}